\documentclass[runningheads]{llncs}
\usepackage[T1]{fontenc}
\usepackage{lmodern}
\usepackage{graphicx}
\usepackage{amsmath}
\usepackage{xcolor}
\usepackage[colorlinks=true,linkcolor=blue!60!black,citecolor=blue!60!black,urlcolor=blue!60!black]{hyperref}
\usepackage{orcidlink}

\graphicspath{{figures/}}
\begin{document}

\title{Accuracy Is Not Enough: A Cross-Architecture Audit of Demographic
Bias in Deep Knowledge Tracing}

\titlerunning{Auditing Demographic Bias in Deep Knowledge Tracing}
\author{Dang Quang Minh\inst{1}\orcidlink{0009-0000-9113-6680}\and
Nguyen Dung Son\inst{1}\orcidlink{0009-0005-6315-9339}\and
Nguyen Huu Loi\inst{2}\orcidlink{0000-0001-7987-0348}\and
Truong Viet Vu\inst{3}\orcidlink{0009-0004-3941-1306} \and
Nguyen Thai Anh\inst{3}\orcidlink{0009-0005-5600-6510}\thanks{Corresponding author.}}
\institute {FPT Polyschool, FPT University, Hanoi, Vietnam \and
FPT Polyschool, FPT University, Can Tho, Vietnam\and
Faculty of Information Technology, Van Lang School of Technology,\\
Van Lang University, Ho Chi Minh City, Vietnam\\
\email{anh.nt@vlu.edu.vn}}
\authorrunning{D. Q. Minh et al.}

\maketitle

\begin{abstract}
Deep knowledge tracing models decide, implicitly, which students an adaptive
learning system believes have mastered a skill, yet nearly everything the
field knows about their demographic fairness comes from studies of Bayesian
knowledge tracing; the deep models that actually power modern systems
have, to our knowledge, received no comparable audit across
architectures. We close this gap with a comprehensive demographic
fairness audit of deep
knowledge tracing: four architectures (DKT, DKVMN, SAKT, AKT)
trained under three regimes (standard, reweighting, adversarial) on two
public datasets with demographic metadata, Eedi (15.9M interactions) and
OULAD (167k after preprocessing), evaluated with ABROCA, student-level
bootstrap confidence intervals, and permutation tests that address recent
critiques of fairness-metric instability. Three findings emerge. \emph{(i)}
Bias is real but context-dependent: every architecture shows a
significant socioeconomic ABROCA on Eedi (0.018 to 0.023, $p < 0.005$),
with per-group AUC consistently lower for economically disadvantaged
students, while gender bias is significant on OULAD for three of the
four architectures after multiplicity correction yet null or negligible
on Eedi. \emph{(ii)} The most accurate architecture
is the most biased: AKT gains about 4 points of AUC from item-level Rasch
embeddings and shows the largest socioeconomic ABROCA, exceeding every
other architecture under a paired student-level bootstrap ($p \leq 0.002$);
ablating only the Rasch embeddings removes the accuracy gain and the
excess bias together, providing controlled ablation evidence linking
the Rasch component to both. \emph{(iii)} Standard
mitigation is unreliable: reweighting and adversarial debiasing leave
ABROCA essentially unchanged in every configuration that preserves
accuracy, even though the adversary is pinned at chance at full reversal
strength, a weak-strength positive control rules out a dead probe, and
the sweep between those strengths offers no usable operating point.
\keywords{Knowledge tracing \and Algorithmic fairness \and ABROCA \and
Learning analytics \and Educational data mining.}
\end{abstract}

\section{Introduction}\label{sec:intro}

Knowledge tracing (KT), the task of predicting whether a student will
answer the next item correctly from their interaction history, sits at the
core of adaptive learning systems: its predictions decide which skills a
student is asked to practice, when practice stops, and what a teacher
dashboard reports as mastery \cite{corbett1994,piech2015}. The field has
moved decisively from the classic Bayesian formulation \cite{corbett1994}
to deep sequence models: recurrent (DKT \cite{piech2015}), memory-augmented
(DKVMN \cite{zhang2017}), and attention-based (SAKT \cite{pandey2019}, AKT
\cite{ghosh2020}) architectures now define the state of the art on public
benchmarks \cite{liu2022}.

If a KT model systematically predicts one demographic group worse than
another, the harm would be invisible and compounding: to the extent that
predictions drive mastery decisions, misestimated mastery routes the
affected students to the wrong practice, and no aggregate accuracy metric
will reveal it. The learning-analytics community has developed both
the conceptual framing \cite{doroudi2019,kizilcec2022,baker2022} and the
measurement instrument for exactly this question: ABROCA, the absolute area
between subgroup ROC curves \cite{gardner2019}, is threshold-independent
and avoids the cancellation problem of AUC differences. Yet the empirical
record is strikingly lopsided. Rigorous KT fairness studies to date audit
\emph{Bayesian} knowledge tracing: Zambrano et al.\ found near-equal BKT
performance across demographic groups \cite{zambrano2024}, Stinar et al.\
found skill-level BKT bias far larger than aggregate bias, across
reading-ability groups, with reweighting essentially ineffective
\cite{stinar2025}, and Barrett et al.\ improved BKT
fairness with time augmentation \cite{barrett2024}. The deep models that
replaced BKT in practice have, to our knowledge, never received a
comparable multi-architecture audit, and recent work shows the audit must
be done carefully: ABROCA distributions are highly skewed and inflate by
chance under small samples and class imbalance \cite{borchers2025}.

\begin{figure}[tbp]
\centering
\includegraphics[width=\textwidth]{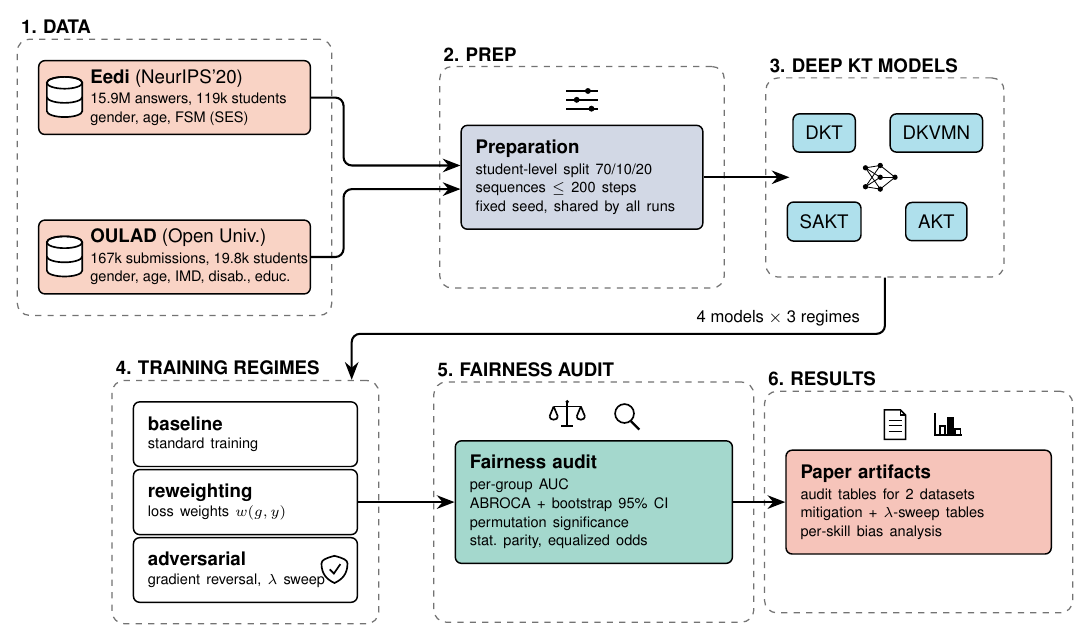}
\caption{\textbf{Study overview.} Two public datasets with demographic
metadata are preprocessed once into fixed student-level splits shared by
every run; four deep KT architectures are trained under three regimes
(baseline, reweighting, adversarial) and audited with per-group AUC, ABROCA
with student-level bootstrap CIs and permutation tests, statistical parity,
and equalized odds. Every run is checkpointed per epoch and resumable on a
single 8\,GB consumer GPU.}
\label{fig:pipeline}
\end{figure}

This paper delivers that audit (Fig.~\ref{fig:pipeline}). We implement
DKT, DKVMN, SAKT, and AKT from scratch with automated causality tests (no
future-response leakage), train each under three regimes (standard,
group-label reweighting \cite{kamiran2012}, and adversarial debiasing with
gradient reversal \cite{ganin2016,zhang2018}), on two standard public
KT-scale datasets with demographic metadata: the Eedi NeurIPS 2020 Education
Challenge data (15.9M answers, 118,971 students, with incompletely
recorded gender, date of birth, and free-school-meal or pupil-premium
eligibility) \cite{wang2020} and OULAD
(166,875 assessment submissions from 19,822 Open University students
after preprocessing, with gender, age, deprivation index, disability, and
prior education) \cite{kuzilek2017}. Every aggregate fairness number
carries a student-level bootstrap confidence interval and a permutation
test, directly answering the interpretability caveats of Borchers and
Baker \cite{borchers2025}; statistical parity \cite{dwork2012} and
equalized odds \cite{hardt2016} are computed for every
configuration.

Our contributions are:
\begin{enumerate}
  \item To our knowledge, the most comprehensive demographic fairness audit
  of deep knowledge tracing to date: 4 architectures $\times$ 3 training
  regimes $\times$ 2 datasets, with multi-seed baselines, an
  adversarial-strength sweep, and significance machinery (bootstrap CIs and
  permutation tests with Benjamini-Hochberg control across the audit grid)
  that prior KT fairness work largely lacks.
  \item Evidence that deep-KT bias is real but context-dependent:
  socioeconomic bias is significant for every architecture on Eedi, gender
  bias is significant on OULAD (for three of four architectures after
  multiplicity correction) but null or negligible on Eedi, and area
  deprivation is null, which cautions against generalizing any single-dataset audit.
  \item The finding that the most accurate architecture amplifies bias:
  AKT's item-level Rasch embeddings buy about 4 AUC points and the largest
  socioeconomic ABROCA at the same time, significant against every other
  architecture under a paired student-level bootstrap and supported by
  a controlled ablation that removes the Rasch embeddings and, with
  them, both the accuracy gain and the excess bias.
  \item A negative result the field needs: two standard mitigation
  techniques, with a mechanism check (a weak-strength positive control
  shows the probe can learn; at $\lambda = 1$ it is pinned at chance),
  fail to
  reduce ABROCA in every accuracy-preserving configuration, and the
  strength sweep offers no usable operating point. Together with a
  practical auditing checklist, this argues that fairness in deployed KT
  must currently be managed by measurement, not by the off-the-shelf
  in-training mitigations we test.
\end{enumerate}

\section{Related Work}\label{sec:related}

\textbf{Knowledge tracing.} Corbett and Anderson's Bayesian knowledge
tracing \cite{corbett1994} dominated intelligent tutoring for two decades
before deep models reframed KT as sequence prediction: DKT with an LSTM
over interaction embeddings \cite{piech2015}, DKVMN with an external
key-value memory over latent concepts \cite{zhang2017}, SAKT with causal
self-attention from the next exercise onto past interactions
\cite{pandey2019}, and AKT with monotonic attention and Rasch-model item
embeddings \cite{ghosh2020}. The pyKT benchmark standardized evaluation and
documented these models' accuracy across public datasets \cite{liu2022}; we
follow its protocol choices (student-level splits, windowed sequences)
while adding the demographic dimension it does not cover.

\textbf{Fairness measurement in education.} Gardner et al.\ introduced
ABROCA and argued for slicing analysis of predictive student models
\cite{gardner2019}; statistical parity and equalized odds transfer from the
general fairness literature \cite{dwork2012,hardt2016}. Surveys of algorithmic bias
in education document harms across the pipeline and call for routine
audits \cite{baker2022,kizilcec2022}. Critically for our design, Borchers
and Baker showed ABROCA point estimates inflate by chance under exactly the
conditions common in education data (small groups, imbalance, modest AUC),
and called for chance-level benchmarking and for methods that establish
the statistical significance of ABROCA differences \cite{borchers2025};
our student-level permutation tests and bootstrap intervals answer that
call.

\textbf{Fairness of knowledge tracing.} The empirical KT fairness
literature is BKT-centric. Doroudi and Brunskill showed in simulation that
KT-driven mastery decisions can be inequitable when the underlying model is
misspecified \cite{doroudi2019}. Zambrano et al.\ audited BKT and
carelessness detectors across demographic groups and found near-equal
performance with no meaningful algorithmic bias \cite{zambrano2024}. Stinar et al.\ audited BKT on a
commercial tutoring system and found aggregate bias small but skill-level
bias large, with reweighting ineffective \cite{stinar2025}; Barrett et al.\
reduced BKT unfairness via time augmentation \cite{barrett2024}. For the
deep models that superseded BKT, we are aware of no comparable
multi-architecture, multi-dataset ABROCA audit with significance testing;
that is the gap this paper fills.

\textbf{Mitigation.} We evaluate the two most portable techniques:
preprocessing-free loss reweighting by group and label \cite{kamiran2012},
as packaged in AIF360 \cite{bellamy2019}, and in-training adversarial
debiasing \cite{zhang2018}, implemented as a gradient-reversal head
\cite{ganin2016} that predicts the sensitive attribute from the model's
hidden state. Both
are architecture-agnostic and thus realistic candidates for a KT vendor;
neither, we find, moves ABROCA. Closest to our setting, FairLISA applies
adversarial fairness filtering to user models (recommendation and
cognitive diagnosis) under partially missing sensitive labels
\cite{zhang2023fairlisa}; it optimizes the fairness of latent user traits,
whereas our question is the audit of next-response prediction across KT
architectures. Training-free post-processing (group-dependent thresholds
\cite{hardt2016}) targets the decision layer directly and is untested
here; we scope our negative result to the two in-training methods
evaluated.

\section{Method}\label{sec:method}

\subsection{Models}
We implement the four canonical deep KT architectures in PyTorch following
the original papers: DKT \cite{piech2015} (LSTM over skill-response
embeddings, per-skill output), DKVMN \cite{zhang2017} (static key and
dynamic value memory with erase-add writes, read-before-write), SAKT
\cite{pandey2019} (the next skill attends causally over past
interactions), and AKT \cite{ghosh2020} (separate question and knowledge
encoders with monotonic exponential-decay attention and Rasch item
difficulty embeddings; we use the standard simplification of raw positional
distance in the decay and a single knowledge-retriever block). DKT, DKVMN,
and SAKT consume skill (knowledge-component) sequences only, and we refer
to them as \emph{skill-level models}; AKT is the only architecture that
additionally embeds item identity, through its Rasch difficulty
parameters. All models
expose the per-step hidden state used for prediction, which the adversarial
regime consumes. Two properties are enforced by automated tests run before
any experiment: every model can memorize a tiny batch (sanity), and no
model's prediction at step $t$ changes when any future response is flipped
(causality; verified exactly, to machine precision, for all four
architectures).

\subsection{Training Regimes}
\textbf{Baseline} is standard training with binary cross-entropy on
next-response prediction. \textbf{Reweighting} rescales the loss of every
prediction with the AIF360-style weight
\[
w(g, y) \propto P(g)\,P(y)\,/\,P(g, y),
\]
computed on the training split, which equalizes the effective weight of
every group-label combination \cite{kamiran2012,bellamy2019}. \textbf{Adversarial} attaches a
two-layer classifier to the per-step hidden features through a
gradient-reversal layer \cite{ganin2016} and trains it jointly to predict
the student's sensitive group \cite{zhang2018}; the reversal (strength
$\lambda$) pushes the KT model toward group-invariant features. We check
the mechanism rather than assume it, in both directions. At $\lambda = 1$
the adversary's final loss sits within 0.005 nats of its chance level,
the entropy of the group base rate (0.536 for FSM on Eedi, not $\ln 2$,
because groups are imbalanced), indicating little group signal recoverable
by the jointly trained probe; at $\lambda = 0.1$ the probe demonstrably
learns on DKT (final loss 0.666 versus the 0.692 student-weighted chance
entropy on OULAD gender; SAKT dips to 0.562 mid-training), ruling out an
implementation defect, though weak-strength learning varies by
architecture and is absent for AKT.
We sweep $\lambda \in \{0.1, 0.3, 1.0\}$ on OULAD.

\subsection{Fairness Measurement}\label{sec:fairmeas}
For each sensitive attribute we compare every eligible group (at least 20
students and 200 test interactions, both classes present) against the
largest group. ABROCA is the integral of the absolute difference between
subgroup ROC curves \cite{gardner2019}. Following the critique of Borchers
and Baker \cite{borchers2025}, every ABROCA value carries (a) a
student-level bootstrap 95\% confidence interval (200 resamples of students
within each group) and (b) a permutation $p$-value (200 permutations of
group labels across the pooled students), so chance-level inflation is
visible rather than hidden. With 200 permutations the smallest attainable
$p$ is $1/201 \approx 0.005$; entries at that value denote the resolution
floor (no permutation reached the observed statistic) and are written
$p < 0.005$. The bootstrap percentiles describe the sampling distribution
of a non-negative statistic and are upward-biased near zero, which is why
significance claims for single-group ABROCA levels rest on the permutation
test, never on a bootstrap interval excluding zero; signed paired
differences (Section~\ref{sec:akt}) do not suffer this bias and are tested
by paired bootstrap. Because the baseline audit spans 44 tests (all
attributes, groups, models, and datasets), we additionally apply
Benjamini-Hochberg correction across the full grid and flag any nominal
finding that does not survive it. We additionally report per-group AUC,
statistical parity difference, and equalized odds difference (the larger
of the absolute TPR and FPR gaps) at threshold 0.5
\cite{dwork2012,hardt2016}.

\section{Experimental Setup}\label{sec:setup}

\textbf{Datasets.} Table~\ref{tab:data} summarizes both datasets after
preprocessing. Eedi \cite{wang2020} provides 15.9M multiple-choice
mathematics answers from 118,971 students with per-student gender, date
of birth, and eligibility for free school meals or pupil premium
(Premium\-Pupil), the dataset's financial-disadvantage flag and a
standard UK poverty proxy, abbreviated FSM here; we order interactions by answer timestamp and use the most
specific subject as the knowledge component. OULAD \cite{kuzilek2017}
provides 166,875 assessment submissions from 19,822 Open University
students with gender, age band, index of multiple deprivation (IMD),
disability, and prior education. Because the official pass mark of 40
\cite{kuzilek2017} labels 95.6 percent of raw submissions correct
(95.9 percent of preprocessed interactions), which is degenerate for
ROC-based auditing, we define correctness as score $\geq 70$ (a stricter
threshold yielding a 70.2 percent positive rate, balanced enough for
ROC-based auditing) and record the threshold in the run metadata. Splits are student-level
70/10/20, fixed once per dataset and shared by every model and regime, so
all comparisons are paired at the student level.
Table~\ref{tab:groups} gives the test-set composition of every audited
group. Two features matter for inference: Eedi groups are large (1,702 to
9,291 students with 117 to 180 predictions each), while OULAD students
contribute only 7 to 8 predictions each, exactly the small-sample regime
in which ABROCA is known to inflate \cite{borchers2025}, which is why
every reported value carries a bootstrap CI and a permutation benchmark.

\begin{table}[tbp]
\centering
\caption{Datasets after preprocessing. Sensitive attributes: gender, age band, IMD band, disability, and highest education for OULAD; gender, age band, and free-school-meal or pupil-premium eligibility (PremiumPupil) for Eedi. Splits are student-level 70/10/20, fixed once and shared by every run.}
\label{tab:data}
\begin{tabular*}{\textwidth}{@{\extracolsep{\fill}}lrrrrr}
\hline
Dataset & Students & Interactions & Items & KCs & Attrs \\
\hline
OULAD & 19,822 & 166,875 & 188 & 188 & 5 \\
Eedi & 118,971 & 15,867,850 & 27,613 & 296 & 3 \\
\hline
\end{tabular*}
\end{table}

\begin{table}[tbp]
\centering
\caption{Test-set composition of every audited group: students, predictions, mean predictions per student, and base rate (fraction correct). Reference groups are marked (ref). Group statistics are properties of the fixed test split and identical for every model and regime.}
\label{tab:groups}
\setlength{\tabcolsep}{4pt}
\begin{tabular*}{\textwidth}{@{\extracolsep{\fill}}llrrrr}
\hline
Dataset & Group & Students & Predictions & Mean len. & Base rate \\
\hline
Eedi & female (ref) & 9,291 & 1,221,615 & 131 & 0.643 \\
 & male & 8,702 & 1,162,465 & 134 & 0.629 \\
 & 16 plus (ref) & 7,587 & 1,109,270 & 146 & 0.616 \\
 & 13 to 15 & 6,564 & 770,568 & 117 & 0.652 \\
 & under 13 & 1,777 & 226,958 & 128 & 0.732 \\
 & non-FSM (ref) & 5,085 & 915,915 & 180 & 0.600 \\
 & FSM & 1,702 & 252,602 & 148 & 0.529 \\
\hline
OULAD & male (ref) & 2,061 & 15,828 & 8 & 0.718 \\
 & female & 1,904 & 13,355 & 7 & 0.694 \\
 & $\leq$35 (ref) & 2,711 & 19,979 & 7 & 0.691 \\
 & $>$35 & 1,254 & 9,204 & 7 & 0.743 \\
 & mid IMD (ref) & 1,591 & 11,636 & 7 & 0.711 \\
 & low IMD & 1,171 & 8,624 & 7 & 0.653 \\
 & high IMD & 1,048 & 7,642 & 7 & 0.749 \\
 & no disability (ref) & 3,638 & 26,771 & 7 & 0.711 \\
 & disability & 327 & 2,412 & 7 & 0.670 \\
 & A level (ref) & 1,784 & 13,436 & 8 & 0.722 \\
 & below A level & 1,477 & 10,462 & 7 & 0.664 \\
 & higher education & 704 & 5,285 & 8 & 0.755 \\
\hline
\end{tabular*}
\end{table}

\textbf{Protocol.} Sequences are windowed at 200 interactions; training
uses Adam, early stopping on validation AUC, and per-epoch checkpointing
(every run resumes exactly after interruption). OULAD baselines are
repeated over three seeds; tables report the primary seed 42 (AUC varies
by at most 0.002 across seeds; cross-seed ABROCA ranges:
Section~\ref{sec:limits}). Eedi runs use a
fixed seed with student-level bootstrap uncertainty. The full grid is 4 architectures $\times$ (1
baseline + 2 mitigations $\times$ 2 target attributes) per dataset, plus
the $\lambda$ sweep and multi-seed repeats: 56 runs in the main program,
72 counting the Rasch ablation, Eedi seed repeats, threshold re-audits,
and a synthetic-data check (planted group bias that the audit pipeline
must and does recover), all on one RTX 2070 Max-Q (8\,GB).

\section{Results}\label{sec:results}

\subsection{Bias Is Real but Context-Dependent}
Table~\ref{tab:eedi} and Table~\ref{tab:oulad} give the baseline audit. On
Eedi, socioeconomic status is the story: every architecture shows a
significant ABROCA for students eligible for free school meals (0.018
to 0.023, all at the permutation floor $p < 0.005$), and the per-group
AUC gaps of $-1.8$ to $-2.3$ points show these students are predicted
worse, while gender is null or negligible (ABROCA $\leq 0.004$). On
OULAD the pattern inverts: gender is significant for every architecture at
the nominal level (ABROCA 0.015 to 0.019, $p \leq 0.025$, with per-group
AUC lower for female students) and survives Benjamini-Hochberg correction across the
44-test grid for three of the four (SAKT, at $p = 0.025$, misses the
cutoff at the primary seed; one to two architectures miss depending on
seed, while nominal gender significance holds for every model in all
three seeds, $p \leq 0.025$), while
area deprivation, disability, and prior education are null for every
architecture. Age effects on OULAD are nominally significant for every
architecture (ABROCA 0.014 to 0.017, $p \leq 0.035$) but none survive the
BH grid; the only age tests that survive multiplicity correction are
AKT's on Eedi (Section~\ref{sec:akt}). The practical
lesson is sharp: a fairness
audit of one dataset, or one attribute, does not license conclusions about
another; the same architecture can be gender-fair on one platform and
gender-biased on another.

\begin{table}[tbp]
\centering
\caption{Baseline audit on Eedi (15.9M interactions; test set 3.1M predictions, 23,795 students). ABROCA against the largest group, student-level bootstrap 95\% CIs, permutation $p$. SES bias is significant for every architecture; gender is null or negligible.}
\label{tab:eedi}
\setlength{\tabcolsep}{2.5pt}
\begin{tabular*}{\textwidth}{@{\extracolsep{\fill}}lccccccc}
\hline
 & & \multicolumn{3}{c}{SES (FSM)} & \multicolumn{3}{c}{Gender} \\
\cline{3-5} \cline{6-8}
Model & AUC & ABROCA & 95\% CI & $p$ & ABROCA & 95\% CI & $p$ \\
\hline
DKT & 0.765 & 0.018 & [0.013, 0.025] & $<$0.005 & 0.002 & [0.001, 0.005] & 0.204 \\
DKVMN & 0.757 & 0.018 & [0.012, 0.025] & $<$0.005 & 0.003 & [0.002, 0.006] & 0.184 \\
SAKT & 0.757 & 0.018 & [0.012, 0.025] & $<$0.005 & 0.003 & [0.002, 0.005] & 0.179 \\
AKT & 0.807 & 0.023 & [0.016, 0.030] & $<$0.005 & 0.004 & [0.002, 0.006] & 0.010 \\
\hline
\end{tabular*}
\end{table}

\begin{table}[tbp]
\centering
\caption{Baseline audit on OULAD (167k interactions; test set 29k predictions, 3,965 students). Gender bias is nominally significant for every architecture and survives Benjamini-Hochberg correction across the audit grid for three of the four (Section~\ref{sec:results}); deprivation (IMD) is null. For near-null effects the bootstrap CI can sit above the point estimate, the small-sample inflation discussed in Section~\ref{sec:fairmeas}; significance rests on the permutation test.}
\label{tab:oulad}
\setlength{\tabcolsep}{2.5pt}
\begin{tabular*}{\textwidth}{@{\extracolsep{\fill}}lccccccc}
\hline
 & & \multicolumn{3}{c}{Gender} & \multicolumn{3}{c}{IMD (low)} \\
\cline{3-5} \cline{6-8}
Model & AUC & ABROCA & 95\% CI & $p$ & ABROCA & 95\% CI & $p$ \\
\hline
DKT & 0.840 & 0.018 & [0.009, 0.031] & $<$0.005 & 0.005 & [0.005, 0.018] & 0.746 \\
DKVMN & 0.841 & 0.018 & [0.007, 0.032] & 0.010 & 0.004 & [0.005, 0.016] & 0.871 \\
SAKT & 0.839 & 0.015 & [0.007, 0.029] & 0.025 & 0.004 & [0.004, 0.017] & 0.905 \\
AKT & 0.839 & 0.019 & [0.008, 0.033] & $<$0.005 & 0.004 & [0.004, 0.017] & 0.910 \\
\hline
\end{tabular*}
\end{table}

Table~\ref{tab:spe} adds the threshold metrics at a 0.5 decision
threshold for the headline attribute of each dataset. Statistical parity
differences look dramatic on Eedi SES (0.110 to 0.116) but conflate model
behavior with the true base-rate gap between the groups (0.529 versus
0.600, Table~\ref{tab:groups}); equalized odds at a single threshold is
nearly identical across architectures (0.089 to 0.092), whereas the
per-group AUC gap separates AKT ($-2.3$ points) from the skill-level
models ($-1.8$). The threshold metrics therefore agree with ABROCA in
direction but cannot resolve the architecture differences that
Section~\ref{sec:akt} shows are systematic, which is the empirical case
for auditing with a threshold-free measure.

\begin{table}[tbp]
\centering
\caption{Threshold-based fairness metrics for the baseline models on the headline attribute of each dataset (decision threshold 0.5): statistical parity difference (SP), equalized odds difference (EO), and the per-group AUC gap (group minus reference). SP partly reflects the true base-rate difference between groups (Table~\ref{tab:groups}). At a single threshold EO is nearly constant across architectures (0.089 to 0.092 on Eedi SES), while the AUC gap ranks AKT worst, illustrating why the audit relies on the threshold-free ABROCA.}
\label{tab:spe}
\setlength{\tabcolsep}{4pt}
\begin{tabular*}{\textwidth}{@{\extracolsep{\fill}}lcccccc}
\hline
 & \multicolumn{3}{c}{Eedi, SES (FSM)} & \multicolumn{3}{c}{OULAD, gender (female)} \\
\cline{2-4} \cline{5-7}
Model & SP & EO & $\Delta$AUC & SP & EO & $\Delta$AUC \\
\hline
DKT & 0.115 & 0.090 & $-0.018$ & 0.018 & 0.010 & $-0.018$ \\
DKVMN & 0.116 & 0.092 & $-0.018$ & 0.015 & 0.008 & $-0.018$ \\
SAKT & 0.114 & 0.090 & $-0.018$ & 0.027 & 0.017 & $-0.015$ \\
AKT & 0.110 & 0.089 & $-0.023$ & 0.023 & 0.016 & $-0.018$ \\
\hline
\end{tabular*}
\end{table}

The label definition is part of that context. Re-running the complete
OULAD baseline audit with correctness thresholds 40 (the official pass
mark, under which 96.1 percent of test predictions are positive) and 60
in place of our primary 70 shows that
the gender effect is threshold-dependent: at 40 it is larger and
nominally significant for every architecture (ABROCA 0.032 to 0.039,
$p \leq 0.045$), at 60 it is uniformly non-significant (0.007 to 0.014,
all $p > 0.05$), and at 70 it is significant again (0.015 to 0.019).
Deprivation flips the same way: null at 70, nominally significant for all
four architectures at 40 (0.032 to 0.039, $p \leq 0.040$). An
auditor who fixed a single operationalization of ``correct'' could have
reported any of three different conclusions from the same raw data. We
therefore treat the threshold as a declared design choice and report it
with the result.

\subsection{The Most Accurate Model Is the Most Biased}\label{sec:akt}
\begin{figure}[tbp]
\centering
\includegraphics[width=0.8\textwidth]{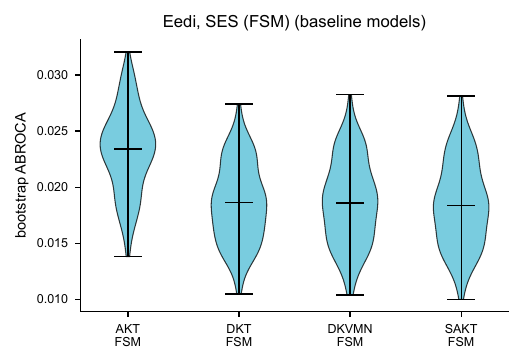}
\caption{\textbf{Bootstrap ABROCA distributions} (200 student-level
resamples) for socioeconomic status on Eedi, baseline models. Distributions
are shown rather than point estimates because ABROCA point estimates
inflate by chance \cite{borchers2025}. The upward shift of AKT's
distribution is visible directly; because the marginal distributions
overlap, the paired analysis in the text is what quantifies the
difference.}
\label{fig:boot}
\end{figure}

AKT dominates accuracy on Eedi (AUC 0.807 versus 0.757 to 0.765), exactly
as its item-level Rasch embeddings predict on a dataset with 27,613
distinct items \cite{ghosh2020,liu2022}. The same embeddings amplify
unfairness: AKT's socioeconomic ABROCA (0.023, CI [0.016, 0.030]) is the
largest of the four, roughly 28 percent above the
skill-level models (0.018). Because all models are evaluated on identical
test students, the comparison can be made paired: bootstrapping students
once per replicate and applying the same resample to both models, AKT's
SES ABROCA exceeds DKT's by $+0.005$ (CI $[+0.0025, +0.0068]$,
$p \leq 0.002$, the two-sided resolution floor at 1,000 replicates;
Fig.~\ref{fig:boot} shows the per-model bootstrap distributions), with
near-identical and
equally significant margins against DKVMN and SAKT. The gap is not
training noise: retraining AKT from three independent seeds moves its
ABROCA by at most 0.0003 (0.0231, 0.0231, 0.0234) and its AUC by less
than 0.0002, and DKT retrained from the same three seeds stays at 0.0180
to 0.0186, so the $+0.005$ separation is stable at both ends.

Two further observations provide strong ablation evidence for the
attribution. First, ablating only the Rasch item-difficulty embeddings
(the AKT-NR variant of \cite{ghosh2020}, identical in every other
respect) simultaneously returns accuracy to skill-level territory (AUC
0.766) and socioeconomic ABROCA to the skill-level value (0.018); the
small gender effect of full AKT (0.004, $p = 0.010$) also becomes
indistinguishable from null (0.003, $p = 0.17$). Second, full AKT shows
the only substantial age reversal: it predicts under-13 students
\emph{better} than the 16-plus reference ($+1.7$ AUC points; ABROCA 0.017
at the permutation floor $p < 0.005$; the skill-level models' gaps are at
most $\pm 0.3$ points), and
the same ablation flips it back to a small gap ($-0.4$ points, close
to the $-0.3$ to $+0.2$ of the skill-level models); one plausible
mechanism, which we do not test directly, is that item-level difficulty
parameters may absorb demographically correlated item exposure. Whatever
AKT buys with its Rasch embeddings, disadvantaged subgroups pay part of
the bill. This is, to our knowledge, the first ablation-verified
demonstration in KT that an architecture upgrade purchased accuracy at a
measurable fairness cost, a tension previously discussed abstractly
\cite{kizilcec2022,baker2022}.

\subsection{Standard Mitigation Does Not Move the Needle}
Table~\ref{tab:mitigation} applies both mitigations to the attribute where
bias is largest on each dataset. The result is a clean negative: ABROCA
changes by at most 0.0015 (exact values; table cells are rounded to three
decimals) in every configuration except the AKT
adversarial run on OULAD, where $\lambda = 1$ collapses AUC from 0.839 to
0.711 to buy near-zero ABROCA. The sweep in Table~\ref{tab:lambda} shows
this is a cliff, not a curve: at $\lambda \leq 0.3$ changes stay within
seed-level noise, and at
$\lambda = 1$ only AKT trades accuracy for fairness catastrophically.
Paired inference sharpens the null. Applying the same student resample to
the mitigated and the baseline run of each architecture on OULAD, we find
reweighting deltas bounded by $\pm 0.0014$ ABROCA with $p \geq 0.28$
everywhere (for the skill-level models the 95\% CIs are as tight as
$\pm 0.0003$) and non-collapsing adversarial deltas of at most $+0.0015$
($p \geq 0.27$).
Even the apparent improvement of the collapsed AKT run ($-0.014$; table
cells are rounded) is not
distinguishable from zero ($p = 0.20$), because a model that has lost 13
AUC points produces noisy group ROCs. These intervals are conditional on
the trained networks; training-seed variability on OULAD gender reaches
about 0.003 ABROCA (Section~\ref{sec:limits}), and the mitigation deltas fall
well within it, reinforcing the null. Mitigation is not merely
ineffective on average; it is ineffective for every architecture
individually. Crucially, the null is not a broken implementation: the adversary sits at
its chance level (base-rate entropy) in every $\lambda = 1$ run while
demonstrably learning at $\lambda = 0.1$ for DKT and SAKT (AKT's probe
stays at chance even at $\lambda = 0.1$, so the capacity check rules out
less there), meaning the hidden
features already carry little group signal recoverable by the probe, and the residual ABROCA plausibly originates
elsewhere (plausibly group-conditional label noise and coverage, which
reweighting and representation invariance cannot repair). This extends the
BKT-era finding of Stinar et al.\ \cite{stinar2025} to the deep regime and
sharpens it with a verified mechanism check.

\begin{table}[tbp]
\centering
\caption{Mitigation targeted at the attribute shown: ABROCA of the target group (and overall AUC) under standard training, reweighting, and adversarial training with gradient reversal ($\lambda = 1$). Mitigation leaves ABROCA essentially unchanged in every configuration except the AKT accuracy collapse on OULAD.}
\label{tab:mitigation}
\setlength{\tabcolsep}{3pt}
\begin{tabular*}{\textwidth}{@{\extracolsep{\fill}}llrrr}
\hline
Setting & Model & Baseline & Reweight & Adversarial \\
\hline
Eedi, SES (FSM) & DKT & 0.018 (0.765) & 0.018 (0.764) & 0.018 (0.759) \\
 & DKVMN & 0.018 (0.757) & 0.018 (0.756) & 0.018 (0.752) \\
 & SAKT & 0.018 (0.757) & 0.018 (0.756) & 0.019 (0.753) \\
 & AKT & 0.023 (0.807) & 0.023 (0.806) & 0.023 (0.800) \\
\hline
OULAD, gender & DKT & 0.018 (0.840) & 0.018 (0.840) & 0.018 (0.838) \\
 & DKVMN & 0.018 (0.841) & 0.018 (0.841) & 0.018 (0.839) \\
 & SAKT & 0.015 (0.839) & 0.015 (0.839) & 0.017 (0.837) \\
 & AKT & 0.019 (0.839) & 0.020 (0.838) & 0.004 (0.711) \\
\hline
\end{tabular*}
\end{table}

\begin{table}[tbp]
\centering
\caption{Adversarial strength sweep on OULAD (target: gender). Cells give gender ABROCA (overall AUC). Adversarial debiasing shows a cliff rather than a smooth tradeoff: weak $\lambda$ leaves bias unchanged, and only AKT at $\lambda = 1$ trades 13 points of AUC for near-zero ABROCA.}
\label{tab:lambda}
\begin{tabular*}{\textwidth}{@{\extracolsep{\fill}}lrrrr}
\hline
Model & Baseline & $\lambda=0.1$ & $\lambda=0.3$ & $\lambda=1$ \\
\hline
DKT & 0.018 (0.840) & 0.018 (0.839) & 0.020 (0.839) & 0.018 (0.838) \\
DKVMN & 0.018 (0.841) & 0.018 (0.840) & 0.018 (0.840) & 0.018 (0.839) \\
SAKT & 0.015 (0.839) & 0.014 (0.838) & 0.015 (0.838) & 0.017 (0.837) \\
AKT & 0.019 (0.839) & 0.018 (0.838) & 0.017 (0.839) & 0.004 (0.711) \\
\hline
\end{tabular*}
\end{table}

\subsection{Where the Bias Lives}
\begin{figure}[tbp]
\centering
\includegraphics[width=0.8\textwidth]{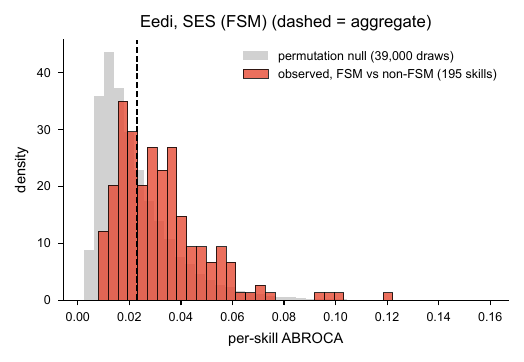}
\caption{\textbf{Per-skill SES bias} on Eedi for AKT: density of
per-skill ABROCA (FSM versus non-FSM, 195 skills with $\geq 100$ test
interactions per group) against the skill-level permutation null (gray),
which quantifies the small-sample inflation of ABROCA. The dashed line
marks the aggregate ABROCA; 39 of 195 skills remain significant after
BH-FDR control, an exploratory upper bound given the anti-conservative
interaction-level permutation null (see text).}
\label{fig:perskill}
\end{figure}

Aggregate ABROCA understates the worst case, but per-skill values must
themselves be calibrated against chance: at the per-skill sample sizes in
our data (131 to 7,555 interactions in the smaller group, median 880),
skill-level permutation nulls (200 label permutations at each
skill's exact sample sizes) already produce a median ABROCA of 0.018, the
small-sample inflation Borchers and Baker warn about \cite{borchers2025}.
Against these nulls, the observed distribution on AKT's Eedi predictions
(195 skills with at least 100 test interactions per SES group) shifts
clearly upward (median 0.029 versus 0.018; 90th percentile 0.055 versus
0.042), 39 of 195 skills remain individually significant after
Benjamini-Hochberg FDR control (an exploratory upper bound; see the
exchangeability caveat below), and the worst skill reaches 0.122, more
than five times the aggregate value of 0.023
(Fig.~\ref{fig:perskill}). One caveat on exchangeability: the per-skill
nulls permute interactions within a skill, not students, which ignores
within-student correlation and makes them mildly anti-conservative, so
the 39-skill count is an exploratory upper bound rather than a confirmed
count; the aggregate tests elsewhere in
the paper all permute at the student level. This mirrors, in the deep regime, the
skill-level concentration Stinar et al.\ reported for BKT
\cite{stinar2025}: a system that looks mildly biased on average can be
severely biased exactly where a particular topic and a particular group
meet. Aggregate-only audits, and uncalibrated per-skill ones, would both
misread it.

\section{Discussion}\label{sec:discussion}

\subsection{A Practical Auditing Checklist}
Our results reduce to five rules for anyone deploying KT. \emph{(1)} Audit
per dataset and per attribute; portability of fairness conclusions is the
exception, not the rule. \emph{(2)} Report ABROCA with student-level
bootstrap intervals and a permutation test; the OULAD age effects,
nominally significant for every architecture, do not survive multiplicity
control, exactly the trap \cite{borchers2025}
warns about. And audit per skill, not only in aggregate: our worst-skill bias is
five times the aggregate value. \emph{(3)} Re-audit after every architecture upgrade: our largest
bias appeared with the most accurate model. \emph{(4)} Do not rely on
reweighting or adversarial training to fix an audit failure; verify the
mechanism, but expect the needle not to move. When an audit fails, the
realistic levers are architecture rollback for the affected deployment,
human review of mastery decisions in the flagged skill-group cells, and
threshold post-processing at the decision layer (untested here). \emph{(5)} Track subgroup
coverage at data collection time; where demographic labels are missing
(71 percent of Eedi students lack SES labels), the audit itself is
partial.

\subsection{Limitations and Ethical Considerations}\label{sec:limits}
Demographic labels are self- or institution-reported and incomplete, and
the missingness may itself be biased; our eligibility thresholds exclude
tiny groups rather than pretending precision. The OULAD correctness
threshold is a design choice we document; the re-audits at
thresholds 40 and 60 (Section~\ref{sec:results}) show the gender
conclusion is sensitive to this choice, a caution we pass on to future
auditors. Eedi mitigation runs use a single seed, but the baseline
seed-repeat experiments bound Eedi training variance at 0.0006 ABROCA
across three seeds of both AKT and DKT, an order of magnitude below the
effects reported. OULAD seed variance is larger (gender ABROCA moves by
up to 0.0034 across the three baseline seeds, and up to 0.005 on other
attributes, comparable to or above the mitigation deltas),
which is why the OULAD mitigation conclusions rest on the paired
student-level bootstrap within a fixed seed, and why gender significance
is checked in every seed (nominal $p \leq 0.025$ for every
architecture). ABROCA magnitudes of 0.02 are small in absolute terms;
we argue they matter because they are systematic, significant, immune to
standard mitigation, and concentrated on protected groups, but we
deliberately do not claim immediate downstream harm without a
decision-layer study. Finally, fairness auditing with sensitive attributes
must itself protect students: both datasets are public and de-identified,
and our audit uses no student-level demographic data beyond what the
datasets already publish.

\section{Conclusion}\label{sec:conclusion}
Across four architectures, three training regimes, and two datasets, deep
knowledge tracing shows demographic bias that is real, context-dependent,
amplified by the most accurate architecture, and untouched by the two
in-training mitigations we test, with the mechanism checked in both
directions. The field's fairness evidence can no
longer stop at Bayesian knowledge tracing: deep KT needs routine,
statistically careful audits of exactly the kind this paper provides,
runnable end-to-end on a single consumer GPU.

\bibliographystyle{splncs04o}
\bibliography{refs}

\end{document}